\documentclass[letterpaper, 10 pt, conference]{ieeeconf}
\IEEEoverridecommandlockouts

\usepackage{graphicx}
\usepackage{cite}
\usepackage{tikz}
\usepackage{amsmath}
\usepackage{mathtools}
\usetikzlibrary{patterns}
\usepackage{bbm}
\usepackage{bm}
\usepackage{amsfonts}

\graphicspath{{images/}}

\title{\LARGE \bf
	Learning Negotiating Behavior Between Cars in Intersections using Deep Q-Learning
}

\author{Tommy Tram$^{1,2}$, Anton Jansson$^{1}$, Robin Gr\"onberg$^{1}$, Mohammad Ali$^{1}$, Jonas Sj\"oberg$^{2}$
	\thanks{This work was partially supported by the Wallenberg AI, Autonomous Systems and Software Program (WASP) funded by the Knut and Alice Wallenberg Foundation.}
	\thanks{$^{1}$Tommy Tram, Anton Jansson, Robin Gr\"onberg and Mohammad Ali are with Zenuity AB, Gothenberg, Sweden
		{\tt\small \{tommy.tram, anton.jansson, robin.gronberg, mohammad.ali\}@zenuity.com}}%
	\thanks{$^{2}$Tommy Tram and Jonas Sj\"oberg are with the Department of Electrical Engineering, Chalmers University of Technology, Gothenberg, Sweden
		{\tt\small \{tram, jonas.sjoberg\}@chalmers.se}}%
}

\begin{document}

\maketitle

\begin{abstract}
This paper concerns automated vehicles negotiating with other vehicles, typically human driven, in crossings with the goal to find a decision algorithm by learning typical behaviors of other vehicles. The vehicle observes distance and speed of vehicles on the intersecting road and use a policy that adapts its speed along its pre-defined trajectory to pass the crossing efficiently.
Deep Q-learning is used on simulated traffic with different predefined driver behaviors and intentions. The results show a policy that is able to cross the intersection avoiding collision with other vehicles 98\%of the time, while at the same time not being too passive. 
Moreover, inferring information over time is important to distinguish between different intentions and is shown by comparing the collision rate between a Deep Recurrent Q-Network  at 0.85\% and a Deep Q-learning at 1.75\%.


\end{abstract}

\IEEEpeerreviewmaketitle

\section{Introduction}
The development of autonomous driving vehicles is fast and there are regularly news and demonstrations of impressive technological progress \cite{Bojarski2016EndCars}. However, one of the largest challenges does not have to do with the autonomous vehicle itself but rather their interaction with human driven vehicles in mixed traffic situations. Human drivers are expected to follow traffic rules strictly, but in addition they also interact with each other in a way which is not captured by the traffic rules  \cite{Liebner2012DriverModel, Lefevre2012EvaluatingIntentions}. This {\em informal} traffic behavior is important, since the traffic rules alone may not always be enough to give the safest behavior. This motivates the development of control algorithms for autonomous vehicles which behave in a "human-like" way, and in this paper we investigate the possibilities to develop such behavior by training on simulated vehicles. 

In \cite{Shalev-ShwartzSafeDriving} they raise two concerns when using Machine learning, specially Reinforcement learning, for autonomous driving applications: ensuring functional safety of the Driving Policy and that the Markov Decision Process model is problematic, because of unpredictable behavior of other drivers.
In the real world, intentions of other drivers are not always deterministic or predefined. Depending on their intention, different actions can be chosen to give the most comfortable and safe passage through an intersection.
They also noted that in the context of autonomous driving, the dynamics of vehicles is Markovian but the behaviors of other road users may not necessarily be Markovian. 

These two concerns are addressed using a Partially Observable Markov Decision Process (POMDP) as a model and Short Term Goals (STG) as actions. With a POMDP the unknown intentions can be estimated using observations and that has shown promising results for other driving scenarios \cite{BrechtelProbabilisticPOMDPs}. The POMDP is solved using a model-free approach called Deep (Recurrent) Q-Learning. 
An initial study of this approach was performed in \cite{masterthesis} and in this paper we show that the policy is able to learn a negotiating behavior without knowing other drivers' intentions. 
With this approach a driving policy can be found using only observations without defining the MDP states. Since we do not train on human driven vehicles, the results presented here cannot be considered  human-like, but the general approach, to train the algorithms using traffic data, is shown working, and  a possible next step could be to start with the pre-tuned policies from this work, and to continue the training in real traffic crossings. 


\section{Overview}
This paper starts by introducing the system architecture and defining the actions, observations and POMDP in Section \ref{sec:model}. The final strategy of what action to take at a given situation is called a policy and is described in Section \ref{sec:policy}. Deep Q-learning is used to find this policy, which uses a neural network to approximate a Q-value and is described in Section \ref{sec:method} together with techniques used to improve the learning, such as Experience replay, Dropout and a recurrent layer called Long Short-Term Memory (LSTM). We then present the simulation, reward function and neural network configurations in Section \ref{sec:implementation}. The results are then presented in Section \ref{sec:results} comparing the effect of the methods mentioned in Section \ref{sec:method}. Finally, the conclusion and brief discussion is presented in Section \ref{sec:conclusion}.

\section{Problem formulation}
\label{sec:model}
The objective is to drive along a main road that has one or two intersections with crossing traffic and control the acceleration in a way that avoids collisions in a comfortable way. All vehicles are assumed to drive along predefined paths on the road where they can either speed up or slow down to avoid collisions in the crossings. 
In this section the system architecture is defined along with the environment, observations and actions. 


\subsection{System architecture}
\begin{figure}[t]
	\centering
	\vspace{0.3cm}
	\includegraphics[width=0.9\columnwidth]{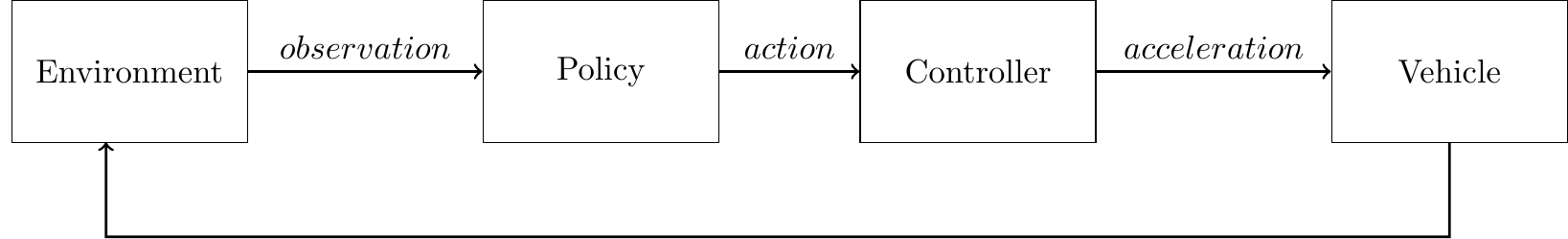}
	\caption{Representation of the architecture}
	\label{fig:Architecture}
	\vspace{-0.3cm}
\end{figure}\
Environment is defined as the world around the ego vehicle, including all vehicles of interest and the shape/type of the intersection. The environment can vary in different ways, e.g. number of vehicles and intersections or the distance to intersections. The environment is defined by the simulation explained in section \ref{sec:simulation}. We assume that the ego vehicle receives observations from this environment at each sampling instant, as shown in Fig. \ref{fig:Architecture}. A policy then takes these observations and chooses a high level action that is defined in more detail in section \ref{sec:observations}. These actions are sent to a controller that calculates the appropriate acceleration request given to the ego vehicle, which will influence the environment and impact how other cars behave. 

\subsection{Actions as Short Term Goals}
\label{sec:actions}
Motivated by the insight that the ego vehicle has to drive before or after other vehicles when passing the intersection, decisions on the velocity profile is modeled by simply keeping a distance to other vehicles until they pass. This is done by defining the actions as Short Term Goals (STG), eg. keep set speed or yield for crossing car. 
This allows the properties of comfort on actuation and safety to be tuned separately, reducing the policy selection to a classification problem. 
The actions are then as follows:
\begin{itemize}
	\item \textit{Keep set speed}: Aims to keep a specified maximum speed $v_{\max}$, using a simple P-controller
	
	\begin{equation}
	a^e_p = K (v_{max} - v^e)
	\label{eq:p_control}
	\end{equation}
	
	where $a^e_p$ is the acceleration request and $v^e$ is the velocity of ego vehicle towards the center of the intersection, while $K$ is a proportional constant. 
	
	\item \textit{Keep distance to vehicle $N$}: Will control the acceleration in a way that keeps a minimum distance to a chosen vehicle $N$, a {\em Target Vehicle}, and can be implemented using a sliding mode controller, where the acceleration request is computed as
	
    \begin{equation}
    	a^e_{sm} = \frac{1}{c_2} (- c_1 x_2 + \mu sign(\sigma(x_1, x_2))) 
    	\label{eq:sliding_mode}
    \end{equation}
    
    $$
    \text{where}
    \begin{cases}
    x_1 = p^t - p^e \\
    x_2 = v^t - v^e
    \end{cases}
    $$

    where $p^e$ and $p^t$ is the position of ego and target vehicle respectively, shown in Fig. \ref{fig:Observations}, and $v^t$ is the velocity of target vehicle. $c_1$ together with $c_2$ are calibration parameters that can be set to achieve wanted performance with a surface
    
    \begin{equation}
    	\sigma = c_1 x_1 + c_2 x_2
    \end{equation}
    
    The final acceleration request is then achieved by
    
    \begin{equation}
		a^e = \min(a^e_{sm}, a^e_p )
		\label{eq:control}
    \end{equation}

    For more detailed information about sliding mode see \cite{SlidingMode}. 
    To distinguish between different cars to follow, each vehicle will have its own action. 
    
    \item \textit{Stop in front of intersection}: Stops the car at the next intersection. Using the same controller as eq. \ref{eq:control} while setting $v^t = 0$ and $p^t$ to start of intersection, the controller can bring ego vehicle to a comfortable stop before the intersection. 
	
\end{itemize}
\subsection{Observations that make up the state}
\label{sec:observations}
\begin{figure}[!h]
	\centering
	\includegraphics[width=0.6\columnwidth]{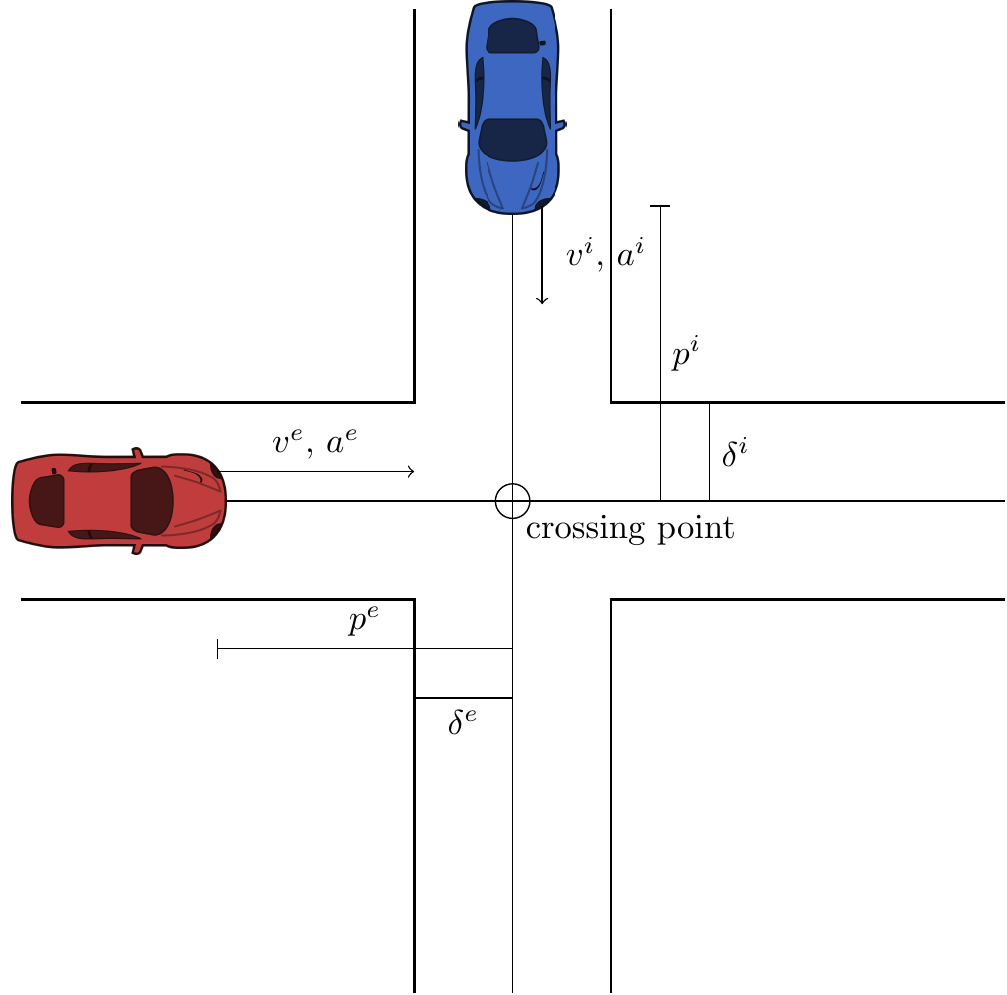}
	\caption{Observations that makes the state}
	\label{fig:Observations}
\end{figure}

%

A human driver is, generally, good at assessing a scenario and it is hard to pin-point what information is used in their assessment. Therefore some assumptions are made on which features that are interesting to observe. The observation $o_t$ at time $t$ is defined as:
\begin{equation}
o_t = [\  p^e_t \quad v^e_t \quad a^e_t \quad \delta^e \quad p^i_t \quad v^i_t \quad a^i_t \quad \delta^i \quad a^{e, A}_{t+1} \  ]^T
\end{equation}

With notations as follows: consider Fig. \ref{fig:Observations}, position of ego $p^e_t$ and other vehicle $p^i_t$ are defined as distance to common reference point, called {\em crossing point}, where $i$ is an index of the other vehicle. 
The start of intersection for ego $\delta^e$ and other vehicle $\delta^i$ also uses the crossing point as reference. These are relevant in case a driver would choose to yield for other vehicles, then they would most likely stop before the start of intersection. The velocity $v^e$ and acceleration $a^e$ of ego vehicle and velocity  $v^i$ and acceleration $a^i$ of the other vehicles are observed to include the dynamics of different actors. The last feature in the observation, $a^{e, A}_{t+1}$, is the ego vehicle's predicted acceleration for each possible action $A$, which can be used to account for comfort in the decision.


\subsection{Partially Observable Markov Decision Processes}
The decision making process in the intersection is modeled as a POMDP. A POMDP works like a Markov Decision Process (MDP) \cite{BellmanMDP} in most aspects, but the full state is not observable. 

At each time instant, an action, $a_t\in \mathcal{A}$, is taken, which will change the environment state $s_t$ to a new state vector $s_{t+1}$. Each action $a_t$ from a state $s_t$ has a value called the reward $r_t$, which is given by a reward function $\mathcal{R}_t $.

One of the unobservable states could be the intentions of other drivers approaching the intersection. The state can only be perceived partially through observations $o_t\in \Omega$ with the probability distribution of receiving observation $o_t$ given an underlying hidden state $s_{t}: o_t \leftarrow \mathcal{O}(s_{t})$, where $\mathcal{O}(s_t)$ is the probability distribution. 

\section{Finding the optimal policy}
\label{sec:policy} 
Assuming the MDP states are not known, we want a model-free method of finding a policy, and for this we use reinforcement learning. The goal is to have an agent learn how to maximize the future reward by taking different actions in a simulated environment. Details on the simulation environment used is described in Section \ref{sec:simulation}. The strategy of which action to take given a state is called a policy $\pi$ and can be modeled in two ways: 

\begin{itemize}
	\item As a stochastic policy $\pi(a|s) = \mathcal{P}[\mathcal{A}=a|\mathcal{S}=s]$
	\item As a deterministic policy $a = \pi(s)$
\end{itemize}
The standard assumption is made that the future reward is discounted by a factor $\gamma$ per time step, making the discounted future reward $\mathcal{R}_t = \sum_{t}^{\tau} \gamma^{t-1} r_t$, where $\tau$ is the time step where the simulation ends, e.g. when the agent crosses an intersection safely. 

Similar to \cite{MnihPlayingLearning}, the optimal action-value function $Q^*(s_t, a_t)$ is defined as the maximum expected reward achievable by following a policy $\pi$ given the state $s_t$ and taking an action $a_t$:

\begin{equation}
Q^*(s_t,a_t)= \max_{\pi} \mathbb{E}[\mathcal{R}_t | s_t, a_t, \pi]
\end{equation}


Using the Bellman equation, $Q^*(s_t, a_t)$ can be defined recursively. If we know $Q^*(s_t, a)$ for all actions $a$ that can be taken in state $s_t$, then the optimal policy will be one that takes the action $a_{t}$ that gives the highest immediate and discounted expected future reward $r_t + \gamma Q^*(s_{t+1}, a_{t+1})$. This gives us: 

\begin{equation}
Q^*(s_t,a_t)= \mathbb{E}[r_t + \gamma \max_{a_{t+1}} Q^*(s_{t+1}, a_{t+1})| s_t, a_t]
\label{eq:q-function}
\end{equation}

The optimal policy $\pi^*$ is then given by taking actions according to an optimal $Q^*(s_t,a_t)$ function: 
\begin{equation}
\pi^*(s_t) = \arg\max_{a_t} Q^*(s_t,a_t)
\label{eq:optimal_policy}
\end{equation}

\section{Method}
\label{sec:method}
In this section we will briefly describe Q-learning and methods used to improve the learning such as, Experience replay, dropout and Long Short-Term Memory.

\subsection{Deep Q-learning}
\label{sec:dqn}
From eq. \ref{eq:optimal_policy}, the optimal policy is defined by taking an action that has the highest expected Q-value. Because the Q-value is not known, a non linear function approximation, such as a neural network, is used to estimate the Q-function. The method is known as Deep Q-Learning \cite{MnihPlayingLearning}.
The neural network used to approximate the $Q$-function is called a Deep Q-network (DQN) and is denoted as $Q(s_t,a_t|\theta^\pi)$, where $\theta^\pi$ is the weight and biases of the neural network for a policy $\pi$. 
The state $s_t$ is the input to the DQN and the output is the $Q$-value for each action $a_t\in \mathcal{A}$.

\subsection{Experience Replay}
Experience replay, as proposed by \cite{MnihPlayingLearning}, is a method that stores all observations $o$, together with taken actions $a$ and their rewards $r$ as an experience memory $E$ and then trains the DQN using sample from experiences $E'$. 

Looking at eq. \ref{eq:optimal_policy}, the optimal policy is greatly affected by Q-function and if the DQN is only trained on recent experiences $E$, the distribution will have a bias towards recent experiences. As shown by \cite{Tsitsiklis1997AnApproximation}, this can give undesired effects on the feedback loops and lead to divergence of the DQN. 

By using experience replay, the network is instead trained on the average of the experience. Thus reducing the time for convergence of the DQN and oscillation due to training on the same experience multiple times \cite{Lin1992Self-ImprovingTeaching}.  

\subsection{Dropout}
Overfitted neural networks have bad generalization performance \cite{Hinton2012ImprovingDetectors} and to help reduce overfitting a technique called dropout was used. 
By temporarily removing some hidden neurons with probability $p$ in the network before each training iteration, the network learns to adapt and generalize instead of depending too strongly on a few hidden neurons. 
For more details, see \cite{Srivastava2014Dropout:Overfitting}.

\subsection{Long short-term memory}
The effect of observed behaviors over time is explored in this paper and is done by adding a recurrent layer to the DQN making it a Deep Recurrent Q-Network (DRQN). A regular recurrent layer has difficulties with longer sequences because of vanishing gradients, and \cite{Hochreiter1997LONGMEMORY} showed that using an LSTM solves this problem.
Instead of storing all information from the previous time sample, LSTM stores information in a memory cell and modifies it by using insert and forget gates. These gates decide if a memory cell should be kept or cleared and is learned by the network. This enables both recent and older observations to be stored and utilized by the network. A sequence length of 4 is used when training the LSTM, where the first 3 observations are only used to build the internal memory state of the LSTM cells, as described in \cite{LamplePlayingLearning}.

\section{Implementation}
\label{sec:implementation}
In this section we go through the experiment implementation. A simulation environment was set up to model the interactions. From section \ref{sec:model}, both the number of observations and actions are dependent on the maximum number of cars. In this paper we consider up to 4 cars. The Deep Q Network can then also be fully defined with the help of observations from section \ref{sec:model} and finally we go through the reward function that defines our behavior. 

\subsection{Simulation environment}
\label{sec:simulation}
The simulation environment is set up as an intersection described in section \ref{sec:observations}. The number of other cars that are observable at the same time can vary from 1-4, while their intentions can vary between an aggressive {\em take way}, passive {\em give way} or a cautious driver. 
The take way driver does not slow down or yield for crossing traffic in an intersection, while the give way driver will always yield for other vehicles before continuing through the intersection. The cautious driver on the other hand, will slow down for crossing traffic but not come down to a full stop. 
With a maximum number of other cars set to 4 all possible actions the ego vehicle can take are:

\vspace{0.2cm}
\begin{itemize}
	\item $\alpha_1$: Keep set speed.
	\item $\alpha_2$: Stop in front of intersection.
	\item $\alpha_3$: Keep distance to vehicle 1.
	\item $\alpha_4$: Keep distance to vehicle 2.
	\item $\alpha_5$: Keep distance to vehicle 3.
	\item $\alpha_6$: Keep distance to vehicle 4.
\end{itemize}
\vspace{0.2cm}

At the start of an episode, the ego vehicle's position and velocity, the number of other vehicles and their intentions are randomly generated. The episode only ends when the ego vehicle fulfills one out of three conditions: 1) Crossing the intersection and reaching the other side, 2) Colliding with another vehicle. or 3) Running out of time $\tau_m$. Each car follows the control law from eq. \ref{eq:control}, trying to keep a set speed while keeping a set distance to the vehicle in front of its own lane.
All cars including the ego vehicle in these scenarios have a maximum acceleration set to $5 m/s^2$, this was set based on comfort and normal driving conditions. 

%
\subsection{Reward function tuning}
Defining the reward function, the distribution was kept around $[-1, 1]$. Large reward values would give large $Q_\pi$-values, so the values are kept small to keep the gradients from growing too large \cite{VanHasseltLearningMagnitude}. 
The reward function is defined as follows:
\begin{align*}
r_t = \hat{r}_t + &\begin{cases}
1 - \frac{\tau}{\tau_m} & \text{on success, }\\
-2                 & \text{on collision}\\
-0.1                & \text{on timeout, i.e. } \tau \ge \tau_m\\
-\left(\frac{j^{e}_t}{j_{\max}}\right)^2\frac{\Delta \tau}{\tau_m}         & \text{on non-terminating updates}
\end{cases} \\
\text{where } \hat{r}_t = &\begin{cases}
-1  & \text{if chosen $a_t$ is not valid}\\
0   & \text{otherwise}
\end{cases}
\end{align*}

\noindent The actions $\alpha_3, \dots, \alpha_6$ described should only be selected when a vehicle is observable and has not crossed the intersection. This is enforced by punishing the agent with a large negative reward $\hat r_t$ if an invalid action was selected. 
Switching between different STG at a high frequency could result in an uncomfortable experience due to high jerk in acceleration. Therefore the agent is also punished for large acceleration jerk $j^{e}_t$, where $\tau$ is the elapsed time since the episode started, $\Delta \tau$ the time between samples and $\tau_m$ is the maximum time before a timeout. 

\subsection{Neural Network Setup}
\begin{figure}[!t]
	\centering
	\includegraphics[width=0.95\columnwidth]{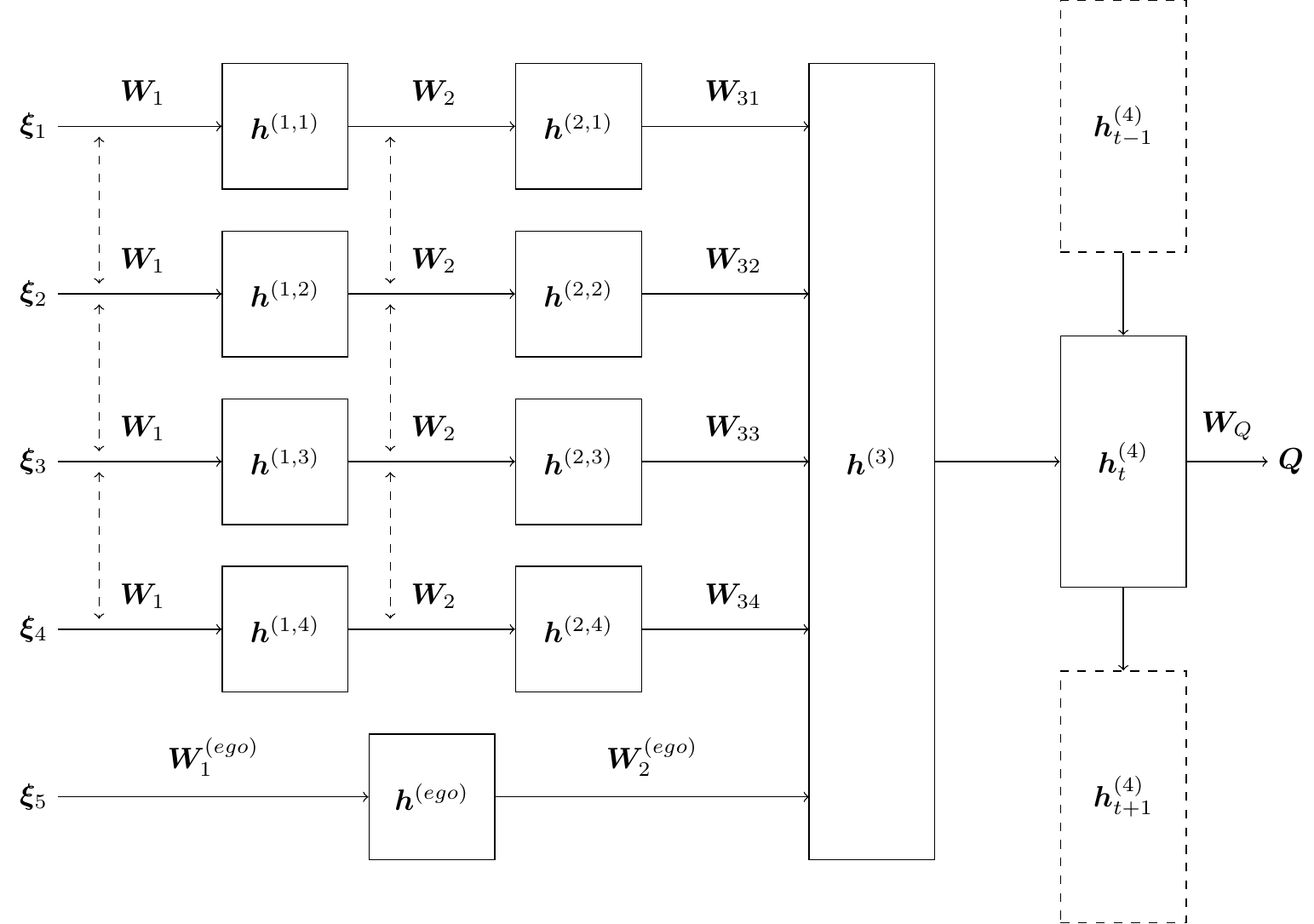}
	\caption{Deep Recurrent Q Network layout with shared weights and a LSTM}
	\label{fig:network}
\end{figure}

The DRQN structure is defined in Fig. \ref{fig:network}. Where $\bm{h}$ are the hidden layers of the network with weights $\bm{W}$. Because the observations $o_t$ from section \ref{fig:Observations}, are used as input to the DRQN, the number of features must be fixed. With up to four other cars, the input vectors $\bm{\xi}$ are as follows:
\vspace{0.3cm}
\begin{itemize}
	\item $ \xi_1 = [\  p^e_t \quad v^e_t \quad a^e_t \quad \delta^e \quad p^1_t \quad v^1_t \quad a^1_t \quad \delta^1 \  ]^T$
	\item $ \xi_2 = [\  p^e_t \quad v^e_t \quad a^e_t \quad \delta^e \quad p^2_t \quad v^2_t \quad a^2_t \quad \delta^2 \  ]^T$
	\item $ \xi_3 = [\  p^e_t \quad v^e_t \quad a^e_t \quad \delta^e \quad p^3_t \quad v^3_t \quad a^3_t \quad \delta^3 \  ]^T$
	\item $ \xi_4 = [\  p^e_t \quad v^e_t \quad a^e_t \quad \delta^e \quad p^4_t \quad v^4_t \quad a^4_t \quad \delta^4 \  ]^T$
	\item $ \xi_5 = [\  a^{e, 1}_{t+1} \quad a^{e, 2}_{t+1} \quad a^{e, 3}_{t+1} \quad a^{e, 4}_{t+1} \quad a^{e, 5}_{t+1} \quad a^{e, 6}_{t+1} \  ]^T$
\end{itemize}
\vspace{0.3cm}
In case a vehicle is not visible, the input vector $\bm{\xi}$ is set to $-\mathbf{1}$, where $-\mathbf{1}$ is a vector of appropriate dimensions with all elements set to one. The maximum speed $v_{\max}$, maximum acceleration $a_{\max}$ and a car's sight range $p_{\max}$ was used to scale all features down to values between $[-1,1]$. 

The output $\bm{Q}$ should be independent of which order other vehicles was observed in the input $\xi_i$. In other words, whether a vehicle was fed into $\xi_1$ or into $\xi_4$, the network should optimally result in the same decision, only based on the features' values. 
The network is therefore structured such that input features of one car, for instance $\xi_1$, are used as input to a sub-network with two layers $\bm{h}^{(1, i)}$ and $\bm{h}^{(2, i)}$. Each other vehicle has a copy of this sub-network, resulting in them sharing weights ($\bm{W}_1$ and $\bm{W}_2$), as shown in Fig. \ref{fig:network}. The first hidden layers are then given by:

\begin{equation}
\bm{h}^{(1, i)} = \tanh\left(\bm{W}_1 \bm\xi_i + \bm{b}_1\right)
\end{equation}
\begin{equation}
\bm{h}^{(2, i)} = \tanh\left(\bm{W}_2 \bm{h}^{(1, i)} + \bm{b}_2\right)
\end{equation}
\begin{equation}
\bm h^{(ego)}   = \tanh\left(\bm{W}^{(ego)}_{1}  \bm\xi_5 + \bm{b}^{(ego)}\right)
\end{equation}

The output of each sub-network, $\bm{h}^{(2, i)}$ and $\bm h^{(ego)}$,  is fed as input into a third hidden layer $\bm{h}^{(3)}$.
The different sub-networks' $\bm{h}^{(2, i)}$ outputs are multiplied with different weights $\bm{W}_{31},\dots,\bm{W}_{34}$ in order to distinguish different cars for different follow car actions. The ego features are also fed into layer 3 with its own weights $\bm{W}^{(ego)}_2$. The neurons in layer $\bm{h}^{(3)}$ combine the inputs by adding them together:
\begin{equation}
\label{eq:shared_weights}
\bm{h}^{(3)} = \tanh\left(\bm{W}^{(ego)}_{2} \, \bm h^{(ego)} + \sum_{i=1}^4 \bm{W}_{3i} \, \bm{h}^{(2, i)} + \bm{b}_3\right)
\end{equation}

The final layer $\bm{h}^{(4)}$ uses the LSTM, described in section \ref{sec:method}. This layer handles the storage and usage of previous observations, making it the recurrent layer of the network. 

\begin{equation}
\bm{h}^{(4)}_t = \text{LSTM}\left( \bm{h}^{(3)} | \bm{h}^{(4)}_{t-1} \right)
\end{equation}

The approximated $\bm{Q}$-value is then
\begin{equation}
\bm{Q} = \bm{W}_Q \bm{h}^{(4)}+ \bm{b}_4
\end{equation}

\section{Results}
\label{sec:results}
Metrics used to evaluate the performance was mainly the success rate followed by collision to timeout ratio (CTR) and average episodic reward. Success rate is defined as number of times the agent reached the end of path without colliding or reaching  the time limit. 
Because both timeouts and collisions are defined as failures, a CTR was used to distinguish the timeouts from collisions for the last 100 episodes where a high value corresponds to more crashes than timeouts. A collision rate corresponding to the total amount of episodes resulting in a collisions is then computed using success rate and CTR averaged. 
To compare the performance between different network structures an average episodic reward is used and is defined as the total reward over an entire episode and averaged over 100 episodes. 
The graphs presented are only using evaluation episodes, with a deterministic policy. For every $300$ training episodes, the policy is evaluated over $300$ evaluation episodes and the evaluation metrics are computed. For more details about the evaluation method see \cite{masterthesis}.

\begin{figure}[!h]
	\centering
	\includegraphics[width=0.7\columnwidth]{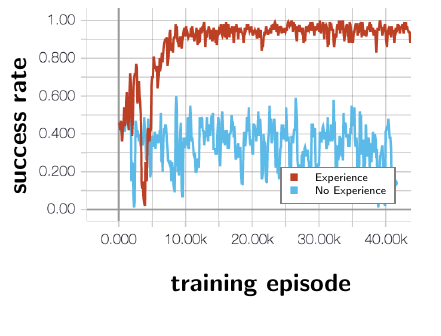}
	\vspace{-0.5cm}
	\caption{Success rate trend comparing using experience replay (red) and not using experience replay (blue)}
	\label{fig:results_experience}
\end{figure}

\begin{figure}[!h]
	\centering
	\includegraphics[width=0.7\columnwidth]{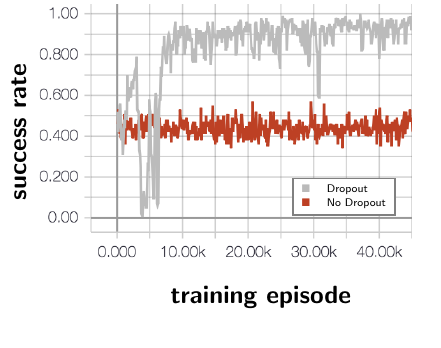}
	\vspace{-0.5cm}
	\caption{Success rate trend comparing using dropout (grey) and not using dropout (red)}
	\label{fig:results_dropout}
\end{figure}

The improvement of using Dropout and Experience replay, from Section \ref{sec:method}, are clearly shown in Fig. \ref{fig:results_experience} and \ref{fig:results_dropout}. Studying the red curve in Fig. \ref{fig:results_experience}, with all methods included, the best policy had a success rate of $98\%$, average episodic reward $0.8$ and CTR at $40\%$. 

\subsection{Effect of using Experience replay and Dropout}
Without either method the success rate does not converge to a value higher than $60\%$. When experience replay was not used, the highest success rate was $53\%$, average episodic reward $-0.1$ and collision to timeout ratio at $90\%$. 

In the case of not using Dropout, not only was the training time significantly higher, the best policy had a average success rate of $58\%$, average episodic reward $-0.7$ and a CTR at $90\%$. Compared to not using dropout, not using experience replay has a higher variation on the success rate.

\subsection{Comparing DQN and DRQN}
\begin{figure}[!h]
	\centering
	\includegraphics[width=0.7\columnwidth]{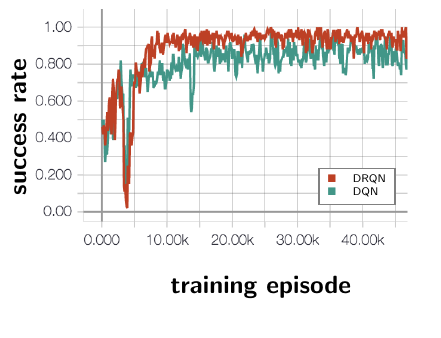}
	\vspace{-0.5cm}
	\caption{Graphs comparing the performance of a DRQN (red) and a normal DQN (green). Showing the DRQN outperforming the DQN.}
	\label{fig:results_recurremt}
\end{figure}
In Fig. \ref{fig:results_recurremt}, we can see the effect off having a recurrent layer, by comparing a DQN without a LSTM layer and with a DRQN with LSTM. The DRQN converges towards a average success rate of $98$\% with a $0.85$\% collision rate while the DQN only reached a success rate of $87.5$\% with a higher collision rate of $1.75$\%.

\subsection{Effect of sharing weights in the network}
\begin{figure}[!ht]
	\centering
	\includegraphics[width=0.8\columnwidth]{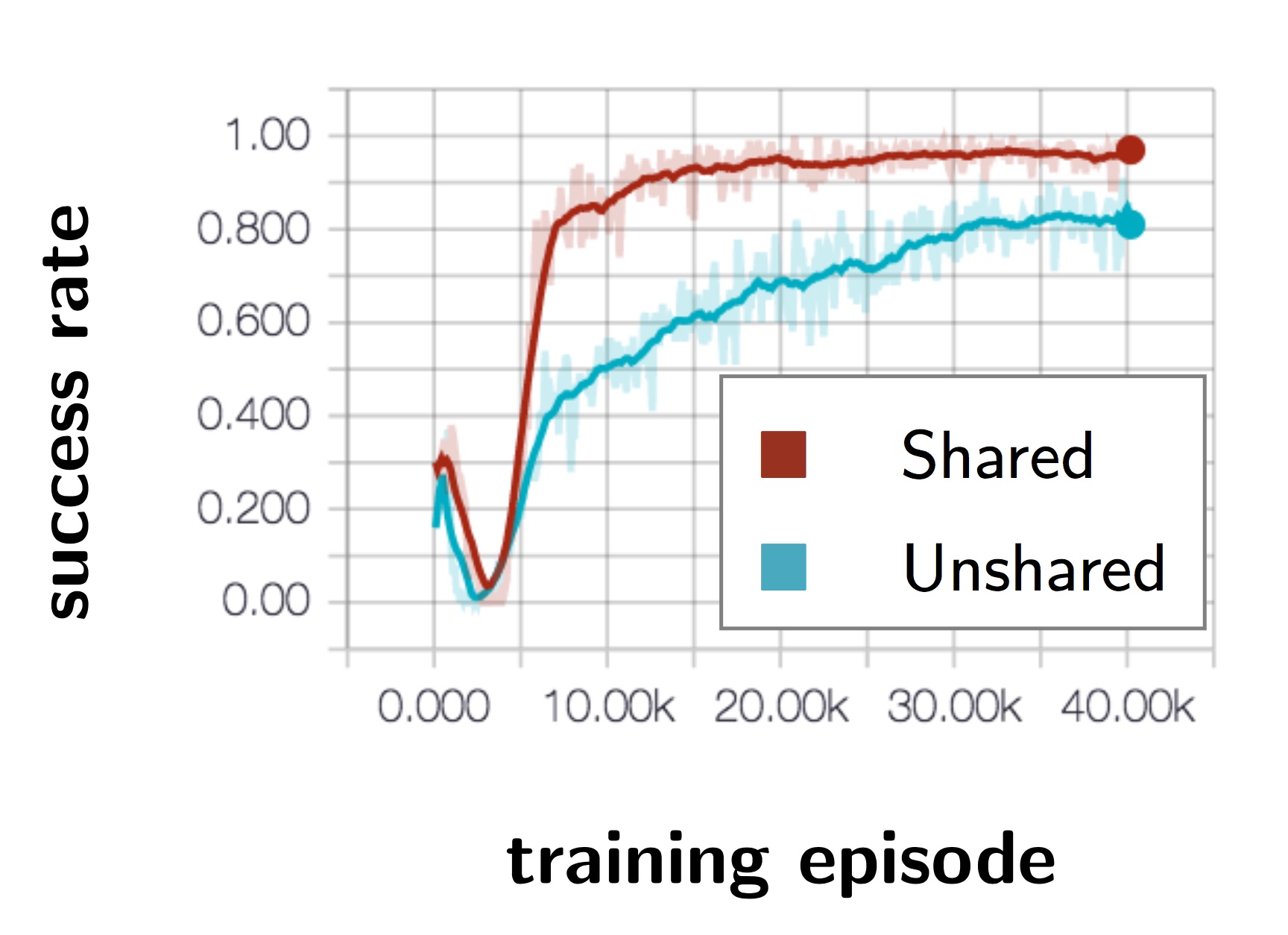}
	\caption{
		Showing the affect of using shared weights for observations from other vehicles. The brown curve represents a network using shared weights while the turquoise curve shows a network without shared weights.}
	\label{fig:results_shared}
\end{figure}
Sharing weights for inputs from other cars showed to converge significantly faster for the DRQN compared to having the network train all weights independently of each other, as shown in Fig. \ref{fig:results_shared}. 

\section{Conclusion}
\label{sec:conclusion}
In this paper, Deep Q-Learning was presented in the domain of autonomous vehicle control. The goal of the ego agent is to drive through an intersection, by adjusting longitudinal acceleration using short-term goals. Short-term goals allowed a smoother and more human-like behavior by controlling the acceleration and comfort with a separate controller. Instead of finding a policy with continuous control output, the problem became a classification problem. 
This resulted in a policy that is able to respond to other vehicles' actions and behaviors without knowing any traffic rules. The policy learned when it is safe to drive ahead of another vehicle or let them pass, without a prediction model as input, while at the same time consider the comfort of the passenger. The trained policy was able to generalize over different types of driver intentions and varied number of cars.

Results show the importance of using a recurrent layer when the environment is modeled as a POMDP. Meaning, the agent needs multiple observations over time in order to better predict some states, e.g. other vehicles' intentions.

Shared weights between observed vehicles in the first layers showed to improve convergence and performance compared to a fully connected network structure. This means that all observed vehicles are processed the same way independently in which order they are fed to the network and in practice would make it easy to scale number of observed vehicles after training. 
These results are limited by simulated traffic scenarios and predefined driving behaviors. For future work we plan on implementing this in a real car and traffic scenarios to record other vehicles' behaviors to improve the policy. 

The success rate of around 98\% is very promising for recognizing behaviors. However, collisions still occur. 
A collision in this paper is defined by two areas overlapping, and in a real world implementation this does not have to mean an actual collision but instead the safety critical area of car. When the two areas overlap, an intervention from a higher safety critical system would intervene. This way, in the low chances a good action could not be found, the safety of the vehicles can still be guaranteed.

In section \ref{sec:actions}, a sliding mode controller was chosen, but this can be replaced by any controller. One other option could be a Model Predictive Controller, where safer actuation can be achieved by using constraints. Also, the actions in this paper used the same controller tuning for all actions, which does not have to be the case. Two action can have the same STG but only differ by the controller's tuning parameters. This way, the agent gains more flexibility while the comfort can maintain intact, possibly increasing the success rate. 

\bibliographystyle{unsrt}
\bibliography{mendeley}


\end{document}